\documentclass{article}

\usepackage{PRIMEarxiv}

\usepackage[utf8]{inputenc} 
\usepackage[T1]{fontenc}    
\usepackage{hyperref}       
\usepackage{url}            
\usepackage{booktabs}       
\usepackage{amsfonts}       
\usepackage{nicefrac}       
\usepackage{microtype}      
\usepackage{lipsum}
\usepackage{fancyhdr}       
\usepackage{graphicx}       
\graphicspath{{media/}}     

\usepackage{amsmath}
\usepackage{orcidlink}
\usepackage{dblfloatfix} 

\newcommand*\N{\mathbb{N}}
\newcommand*\R{\mathbb{R}}
\newcommand*\X{\mathcal{X}}
\newcommand*\Y{\mathcal{Y}}
\newcommand*\E{\mathbb{E}}

\title{
Quantifying Robustness: A Benchmarking Framework for Deep Learning Forecasting in Cyber-Physical Systems
}

\author{
  Alexander Windmann\orcidlink{0000-0002-6522-4262} \\
  Institute of Artificial Intelligence \\
  Helmut Schmidt University \\
  Hamburg, Germany\\
  \texttt{alexander.windmann@hsu-hh.de} \\
   \And
  Henrik Steude\orcidlink{0000-0002-7812-4279} \\
  Institute of Artificial Intelligence \\
  Helmut Schmidt University \\
  Hamburg, Germany\\
  \texttt{henrik.steude@hsu-hh.de} \\
  \AND
  Daniel Boschmann\orcidlink{0009-0003-3345-5351} \\
  Institute of Artificial Intelligence \\
  Helmut Schmidt University \\
  Hamburg, Germany\\
  \texttt{daniel.boschmann@hsu-hh.de} \\
  \And
  Oliver Niggemann\orcidlink{0000-0001-8747-3596}\\
  Institute of Artificial Intelligence \\
  Helmut Schmidt University \\
  Hamburg, Germany\\
  \texttt{oliver.niggemann@hsu-hh.de} \\
}

\begin{document}
\maketitle

\begin{abstract}
Cyber-Physical Systems (CPS) in domains such as manufacturing and energy distribution generate complex time series data crucial for Prognostics and Health Management (PHM).
While Deep Learning (DL) methods have demonstrated strong forecasting capabilities, their adoption in industrial CPS remains limited due insufficient robustness.
Existing robustness evaluations primarily focus on formal verification or adversarial perturbations, inadequately representing the complexities encountered in real-world CPS scenarios.
To address this, we introduce a practical robustness definition grounded in distributional robustness, explicitly tailored to industrial CPS, and propose a systematic framework for robustness evaluation.
Our framework simulates realistic disturbances, such as sensor drift, noise and irregular sampling, enabling thorough robustness analyses of forecasting models on real-world CPS datasets.
The robustness definition provides a standardized score to quantify and compare model performance across diverse datasets, assisting in informed model selection and architecture design.
Through extensive empirical studies evaluating prominent DL architectures (including recurrent, convolutional, attention-based, modular, and structured state-space models) we demonstrate the applicability and effectiveness of our approach. 
We publicly release our robustness benchmark to encourage further research and reproducibility.
\end{abstract}


\begin{figure}[!b]
\fbox{
  \parbox{\dimexpr\linewidth-2\fboxsep-12\fboxrule\relax}{
    \tiny © 2025 IEEE. Personal use of this material is permitted. Permission from IEEE must be obtained for all other uses, in any current or future media, including reprinting/republishing this material for advertising or promotional purposes, creating new collective works, for resale or redistribution to servers or lists, or reuse of any copyrighted component of this work in other works. DOI: \href{https://doi.org/10.1109/ETFA65518.2025.11205527}{10.1109/ETFA65518.2025.11205527}
  }
}
\end{figure}

\section{Introduction}
\label{sec:intro}

Cyber-Physical Systems (CPS) continuously generate large-scale and complex time series data across industrial domains such as manufacturing, process engineering, and energy distribution, which enable vital tasks including Prognostics and Health Management (PHM) \cite{leePrognosticsHealthManagement2014}.
Deep Learning (DL) has proven effective in modeling complex data, yet despite its predictive capabilities the adoption of DL in industrial CPS remains limited due to concerns over robustness \cite{windmannArtificialIntelligenceIndustry2024}.
Robustness, in this context, refers to the model's ability to maintain reliable performance when encountering erroneous or unforeseen inputs \cite{ISOIECTR2021}.
Regulatory developments, such as the European Union’s AI Act, have further highlighted the importance of robust DL systems, urging systematic methodologies for robustness assessment \cite{RegulationEU20242024}.
However, consensus on practical robustness testing methodologies remains elusive \cite{schuettRiskManagementArtificial2024}.
While formal verification methods and adversarial robustness
studies have been prominent in the literature, their practical relevance to industrial scenarios is limited due to unrealistic assumptions \cite{perez-cerrolazaArtificialIntelligenceSafetyCritical2023}.
Recent research emphasizes evaluating model robustness against realistic data quality issues and distributional shifts commonly encountered in real-world settings \cite{zhouRobustnessTestingData2022,dixMeasuringRobustnessML2023,windmannRobustnessGeneralizationPerformance2023,schmidtAssessingRobustnessDataDriven2025}.
However, existing studies stop short of combining an operational robustness metric and a large-scale evaluation on real CPS data, leaving practitioners without a practical benchmark.

Addressing these limitations, we propose a novel, practical robustness definition tailored specifically to industrial CPS time series forecasting.
This definition integrates the principles of distributional robustness by explicitly quantifying model performance degradation under realistic disturbances such as sensor drift, measurement noise, and irregular sampling, which are frequently encountered in real operational environments.
Accompanying this definition, we introduce a comprehensive, systematic robustness testing framework designed for assessing and benchmarking diverse DL architectures, including recurrent networks (LSTM, GRU), attention-based models (Transformers, Informers), convolutional models (TCN), modular architectures (RIMs) and structured state-space models (Mamba), on multiple real-world CPS datasets.

Our key contributions are:
\begin{itemize}
\item We provide a clear, practical definition of robustness of DL models specifically suitable for industrial CPS.
\item We develop a systematic robustness testing framework that quantifies the robustness, bridging theoretical robustness evaluation and real-world applicability.
\item We provide empirical robustness evaluations of DL forecasting models across multiple real-world CPS datasets.
\end{itemize}
Figure~\ref{fig:overview} illustrates the workflow of our proposed robustness testing methodology.
The proposed framework is publicly available at: \url{https://github.com/awindmann/cps-robustness-benchmark}

\begin{figure*}[b!]
    \centering
    \includegraphics[width=.99\textwidth]{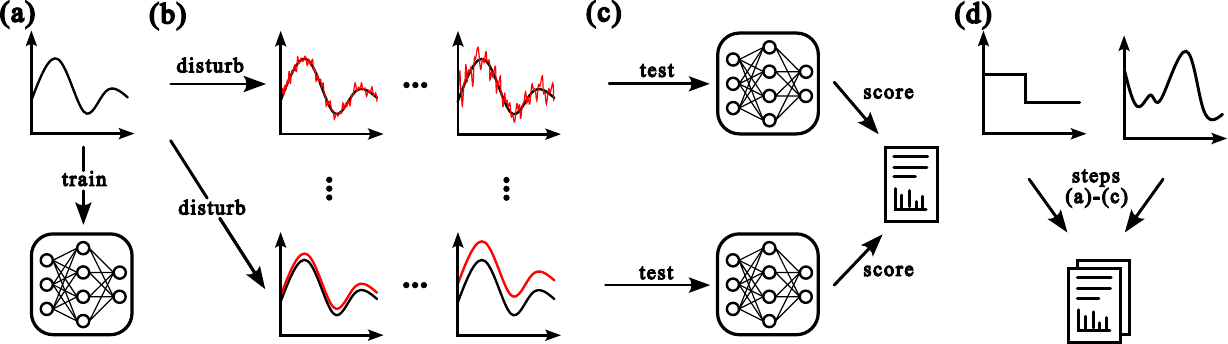}
    \caption{Overview of the proposed robustness evaluation framework. (a) Standardized time series data is partitioned into training, validation, and test subsets, and models are trained accordingly. (b) Realistic disturbances, such as sensor drift, noise, and irregular sampling, are systematically applied to the standardized test subset at varying severity levels. (c) Model predictions are assessed on disturbed test subsets with different severities, and results are used to calculate a robustness score. (d) To ensure comprehensive evaluation, the entire process (a-c) is repeated across multiple datasets for various DL architectures.}
    \label{fig:overview}
\end{figure*}

\section{Related Work}
\label{sec:related_work}

\textbf{Robustness in DL}: Robustness of a DL model is generally described as the ability "to maintain its level of performance under any circumstances" \cite{ISOIECTR2021}, or its capability "to cope with erroneous, noisy, unknown, and adversarial input data" \cite{DINSPEC9200112019}.
It is closely related to other concepts such as reliability \cite{billintonReliabilityEvaluationEngineering1992}, resilience \cite{moskalenkoResilienceResilientSystems2023}, safety \cite{perez-cerrolazaArtificialIntelligenceSafetyCritical2023} and stability \cite{runggerNotionRobustnessCyberPhysical2016}.
DL models are known to be highly sensitive to small input perturbations\cite{szegedyIntriguingPropertiesNeural2014}.
Consequently, much research has focused on adversarial robustness, evaluating model resilience to these small, intentionally crafted perturbations \cite{bastaniMeasuringNeuralNet2016}, leading to formal guarantees and metrics to measure this type of robustness \cite{sinhaCertifyingDistributionalRobustness2018}.
However, adversarial perturbations rarely occur naturally, thus recent research emphasizes measuring robustness to broader, more realistic distributional shifts \cite{buzhinskyMetricsMethodsRobustness2021,hendrycksUnsolvedProblemsML2022}.
Traditional adversarial robustness metrics often fail to sufficiently capture these distributional shifts \cite{jiangFantasticGeneralizationMeasures2019,dziugaiteSearchRobustMeasures2020}.
To address this limitation, recent studies propose evaluating robustness either by assessing performance across multiple distinct environments \cite{yuanOutDistributionGeneralization2022}, by applying broad yet realistic natural distribution shifts (e.g., image blur or style changes) \cite{hendrycksManyFacesRobustness2021}, or by specifically investigating the impact of data-quality issues \cite{schelterJENGAFrameworkStudy2021, mohammedEffectsDataQuality2024}.

\textbf{Robustness for Time Series and CPS}:
Robustness in time series forecasting is generally assessed by evaluating how model performance deteriorates in response to anomalies, commonly measured by increased forecasting errors \cite{chengRobustTSFTheoryDesign2024}, or by measuring the effect on the output distribution, e.g. by measuring the Wasserstein distance \cite{yoonRobustProbabilisticTime2022}.
In CPS, research of robustness has traditionally focused on formal verification \cite{mitschModelPlexVerifiedRuntime2016} or stability in control engineering \cite{runggerNotionRobustnessCyberPhysical2016}.
Other CPS-focused robustness research has explored adversarial attacks \cite{zhouRobustnessTestingData2022}, false-positive predictions under varying grades of handicap \cite{pfrommerReduceHandicapPerformance2023}, and model robustness to added random noise \cite{schmidtAssessingRobustnessDataDriven2025}.
Recent studies have specifically investigated the effects of data quality issues such as sensor drift, outliers and missing values on simulated CPS data \cite{dixMeasuringRobustnessML2023, windmannRobustnessGeneralizationPerformance2023}.

\textbf{Prognostics in CPS}: In CPS contexts, PHM assesses system health and predicts future system behavior \cite{leePrognosticsHealthManagement2014}. 
Many PHM applications rely on DL forecasting models, either by using them to directly predict the remaining useful lifetime (RUL) or by first forecasting health indicators of the system. 
In general, more accurate forecasts of health indicators are strongly associated with improved RUL prediction accuracy \cite{wangHealthIndicatorForecasting2020, sunPredictingFutureCapacity2024}.
Commonly used architectures include recurrent neural networks (e.g., LSTM \cite{hochreiterLongShortTermMemory1997}, GRU \cite{choLearningPhraseRepresentations2014}), modular approaches like Recurrent Independent Mechanisms (RIMs) \cite{goyalRecurrentIndependentMechanisms2021}, Temporal Convolutional Networks (TCN) \cite{leaTemporalConvolutionalNetworks2017}, attention-based models such as Transformers \cite{vaswaniAttentionAllYou2017} and Informers \cite{zhouInformerEfficientTransformer2021}, and structured state-space models like Mamba \cite{guMambaLinearTimeSequence2024}.
Recent findings indicate that even simple feed-forward neural networks can be competitive \cite{zengAreTransformersEffective2023}.
These models represent diverse architectural approaches commonly adopted in CPS forecasting tasks.
Despite their widespread application, the robustness of these architectures has primarily been evaluated using synthetic or idealized datasets, potentially masking their real-world vulnerabilities.

In summary, no prior paper combines an operational robustness score and a large-scale evaluation on real CPS data under realistic disturbances.
Papers that introduce formal definitions validate them only on synthetic case studies \cite{schmidtAssessingRobustnessDataDriven2025,pfrommerReduceHandicapPerformance2023} or focus solely on adversarial noise rather than real-world scenarios  \cite{zhouRobustnessTestingData2022}.
Studies that do examine realistic data-quality issues limit their evaluations to simulations and report robustness as simple performance deltas \cite{dixMeasuringRobustnessML2023,windmannRobustnessGeneralizationPerformance2023}.
This paper closes these gaps by proposing a distribution-grounded robustness score and benchmarking nine forecasting models under representative sensor-fault scenarios across six industrial CPS data sets.

\section{Defining Robustness for CPS}
\label{sec:definition}

This definition will form the foundation for the testing framework introduced in Section~\ref{sec:Framework}.

A CPS monitors several sensors over time, producing multivariate time series data.
We denote this time series as $\mathcal{T} = (\textbf{x}_i)_{i \in \N}$, where each $\textbf{x}_i \in \R^n$ represents the sensor readings at time step $i$. 
Let $\textbf{X} \subset \R^{(t+t') \times n}$ be a dataset of samples extracted from $\mathcal{T}$.
Each sample is a time window containing $t+t'$ consecutive sensor readings, with $t,t' \in \N$ and starting from a randomly chosen time step $i$, and is defined as:
\begin{equation}
    \label{eq:sample(x,y)}
(X,Y)= ((\textbf{x}_{i},  \dots, \textbf{x}_{i+t-1}), (\textbf{x}_{i+t}, \dots, \textbf{x}_{i+t'-1})) \in \textbf{X} .   
\end{equation}
In forecasting, the goal is to use the first $t$ sensor readings 
$X$ to predict the subsequent $t'$ readings $Y$.

We train a model $f \in F$ from the set of prognostic models 
\begin{equation}
    \label{eq:F}
    F = \{f: \R^{t \times n} \to \R^{t'\! \times n}\}
\end{equation} 
on the dataset $\textbf{X}$ to approximate the target $Y$ based on the input $X$. 
The model’s prediction $f(X) = \hat{Y}$ is compared to the actual target $Y$ using a performance measure or loss function 
\begin{equation}
    \label{eq:mu}
\mu: \R^{t'\! \times n} \times \R^{t'\! \times n} \to \R_{\ge 0}.
\end{equation} 
For example, a common choice for $\mu$ is the Mean Squared Error (MSE), defined by
$$
\text{MSE}\,(Y,\hat{Y})=\frac{1}{t'n}\sum_{i=1}^{t'}\sum_{j=1}^{n} \bigl(Y_{ij}-\hat{Y}_{ij}\bigr)^{2}.
$$
We assume that the performance measure is non-negative, with a value of 0 indicating optimal performance.
Since the models are evaluated on a finite dataset, the loss $\mu$ is finite and thus integrable.
If a performance measure does not naturally meet these criteria, it can be transformed via monotonic transformations.

To better model the forecasting setup, we treat each sample $(X, Y)$ (see Eq.~\eqref{eq:sample(x,y)}) as a realization of a random variable $(\X, \Y)$, where $\X: \Omega \to \R^{t \times n}$ and $\Y: \Omega \to \R^{t'\! \times n}$ for a probability space $(\Omega, \mathcal{A}, P)$. 
Using this notation, the goal of forecasting is defined as minimizing the expected loss
\begin{equation*}
    \label{eq:min-exp-loss}
    \min_{f \in F} \, \E_{}\left[\mu(f(\X), \Y) \right].
\end{equation*}
In other words, we are looking for the prognostic model $f\in F$ that can best predict the values of $Y$ based on the preceding time steps $X$.

\begin{figure}[!b]
    \centering
    \includegraphics[width=.7\columnwidth]{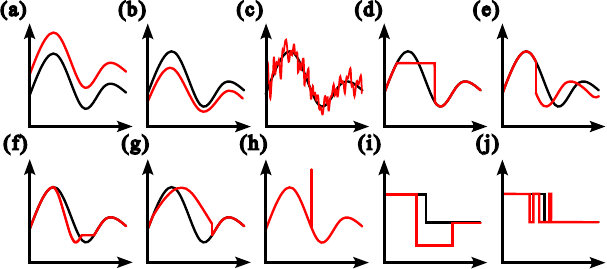}
    \caption{Overview of the disturbance scenarios applied to a subset of sensors of the standardized time series data.
    The black line indicates the original time series, the red line signifies the time series after the disturbance has been applied.
    (a)~\textbf{Drift}: A constant offset is added to the signal. 
    (b)~\textbf{DyingSignal}: The amplitude diminishes by multiplying the values by a factor smaller than 1. 
    (c)~\textbf{Noise}: Gaussian noise is added to the signal. 
    (d)~\textbf{FlatSensor}: The sensor remains constant for a designated time interval. 
    (e)~\textbf{MissingData}: An entire segment of the time series is removed to simulate missing values. 
    (f)~\textbf{FasterSampling}: The sampling rate is temporarily increased, followed by a flat sensor period until the time steps realign. 
    (g)~\textbf{SlowerSampling}: The sampling rate is temporarily decreased until the time steps realign. 
    (h)~\textbf{Outlier}: The signal exhibits an unusually high spike value at a specific time step. 
    (i)~\textbf{WrongDiscreteValue}: A discrete sensor is set to an invalid state for a certain period.
    (j)~\textbf{OscillatingSensor}: A discrete sensor oscillates between two valid states for a certain duration. }
    \label{fig:scenarios}
\end{figure}

However, this formulation assumes stationarity in the underlying data distribution, which typically does not hold in real-world applications due to changing operational conditions, system wear, or unforeseen disturbances.
Consequently, models often experience performance degradation after deployment. 
To assess and improve model robustness realistically, we propose applying disturbances that simulate diverse operating conditions encountered in practice. Specifically, we define a disturbance function $d \in \mathcal{D}$ as
\begin{equation}
\label{eq:disturbance-fct}
    d: [0,1] \times \R^{t \times n} \times \R^{t' \! \times n} \to \R^{t \times n} \times \R^{t' \! \times n},
\end{equation}
which transforms a input-output sample $(X, Y)$ into a disturbed version of itself.
We denote the disturbed sample by
\begin{equation*}
    \label{eq:x_disturbed}
    \left(X^{(s)}_d, Y^{(s)}_d\right):=d(s,X,Y),
\end{equation*}
where the parameter $s\in[0,1]$ controls the severity of the disturbance. 
When $s=0$, no disturbance is applied; when $s=1$, the disturbance reaches its maximum realistic intensity.
In general, as $s$ increases, we expect the disturbed sample to become less similar to the original.
For example, in the case of the {Noise} disturbance, we add Gaussian noise scaled by $s$ to each affected sensor in $X$, making the sample progressively less like the original.
This results in the disturbance function
$$
d_{\text{noise}}\!: (s,X,Y) \mapsto (X + s\mathcal{Z}, Y),
$$ 
where $\mathcal{Z} \sim \mathcal{N}(0, \mathbb{I})$ represents Gaussian noise.
Figure~\ref{fig:scenarios} shows an overview of the disturbances used in this study.

To assess model robustness, we compare performance on disturbed samples with performance on the original samples.
Since the ideal performance measure may vary across applications, we adopt the general performance measure $\mu$ as defined in Eq.~\eqref{eq:mu}.
For a model $f$ (see Eq.~\ref{eq:F}) and a disturbance function $d$ (see Eq.~\ref{eq:disturbance-fct}), we define the relative performance as 
\begin{equation}
\label{eq:mu-rel}
    \mu_\text{rel}(f,d,s,X,Y) = \frac{\mu (f(X),Y) + \varepsilon} { \mu \left(f (X_d^{(s)}),Y_d^{(s)} \right) + \varepsilon},
\end{equation}
where $0 < \varepsilon <\!< 1 $ is a small constant added for numerical stability.
The relative performance $\mu_{rel}$ measures the effect of a disturbance by comparing the performance on the disturbed sample against the original performance.
Note that $\mu_\text{rel}$ makes no statement about the inherent performance of the model, it only measures the effect of the disturbances.

We then define the disturbance robustness score for a model $f$ (see Eq.~\eqref{eq:F}) and a disturbance function $d$ (see Eq.~\eqref{eq:disturbance-fct}) using the relative performance $\mu_\text{rel}$ (see Eq.~\eqref{eq:mu-rel}) as
\begin{equation}
    \label{eq:R-score-dist}
    R_d(f, \mu) = \E_{}\left[\int \limits_0^1 \mu_\text{rel}(f,d,s,\X,\Y)ds\right].
\end{equation}
In other words, for a given disturbance $d$, we gradually increase its severity $s$ from 0 to 1 and integrate the relative performance over this range.
This score $R_d$ reflects how much the disturbance affects the model's performance.
Values of $R_d$ at or around 1 indicate that the disturbance has little to no negative impact, while values close to 0 suggest that the model's performance deteriorates sharply.
Note that while in theory the data could be sampled from the underlying distribution $P_{}$, in practice we chose to sample from the test partition of the dataset.

Finally, to evaluate the overall robustness of a model $f$ (see Eq.~\eqref{eq:F}) using a performance measure $\mu$ (see Eq.~\eqref{eq:mu}), we multiply the disturbance robustness scores $R_d$ (see Eq.~\eqref{eq:R-score-dist}) across the disturbance functions $d$ (see Eq.~\eqref{eq:disturbance-fct}):
\begin{equation}
    \label{eq:R-score-global}
    R(f, \mu) = \prod_{d\in \mathcal{D}} R_d(f, \mu).
\end{equation}
The multiplication ensures that a complete failure in any one disturbance scenario (where $R_d$ is near 0) significantly lowers the overall robustness score, while poor performance across multiple disturbances compounds the effect, providing a nuanced picture of the model's overall robustness.
Alternatively, one could average the scores $R_d$ over all $d\in \mathcal{D}$, but this approach might dilute the impact of a complete failure in one scenario.
Another approach would be to pick the minimum disturbance robustness score, which would highlight the worst-case performance but could ignore more gradual degradations in other scenarios.

\section{Proposed Framework}
\label{sec:Framework}

To evaluate the proposed definition of robustness, we implemented a framework that simulates disturbances in CPS datasets to quantify the robustness of popular DL model architectures.
We focus on a forecasting task as a representative objective, since it requires a deep understanding of the underlying system, which can translate into more precise RUL estimates \cite{wangHealthIndicatorForecasting2020, sunPredictingFutureCapacity2024}. 
The framework is \hyperlink{https://github.com/awindmann/robust-AI-validation}{openly accessible}, supports custom time series data in CSV or Parquet format and can easily integrate models implemented in Python.
An overview of the proposed framework is shown in Figure~\ref{fig:overview}.

First, we preprocess a given multivariate time-series dataset by removing missing values and splitting the data into training, validation, and test sets along three disjunct continuous time segments.
To mitigate information leakage, we purge a small fraction of the data near the boundaries between these segments, ensuring that the model does not see validation or test data during training.
We then standardize all datasets based on the statistics of the training split, which helps during training and ensures that the disturbances are applied consistently.
The final training, validation, and test sets are created by randomly sampling time frames from their respective segments, after which the model is trained on the resulting training data.

The disturbances (see Eq.~\eqref{eq:disturbance-fct}) simulate behavior commonly encountered in real-world CPS environments, like sensor drift, noise and irregular sampling, in gradually increasing severity.
Thus, the robustness score (see Eq.~\eqref{eq:R-score-global}) can effectively measure the effect of such critical scenarios, providing a relevant quantification of model robustness.
Ideally, the severity increment should be infinitesimally small so that the discrete summation closely approximates the integral in Eq.~\eqref{eq:R-score-dist}.
In practice, we approximate the integral with uniform 1\% severity steps, which we found to balance numerical accuracy and computational cost.
Figure~\ref{fig:scenarios} provides an overview of these test scenarios.
For each dataset, we affect a subset of 10\% of the sensors.
Since the dataset is standardized, the effect of each disturbance is comparable across different datasets.
Some disturbances apply solely to continuous signals (e.g., Noise), while others are specific to discrete signals or actuators (WrongDiscreteValue, OscillatingSensor).
Most disturbances are not applied to the portion of the time series to be forecasted, as it would be infeasible or undesirable to predict faulty sensor values.
However, MissingData requires the model to ignore the missing values and forecast using the most recent available time steps, so the section of the time series to predict changes.

The final robustness score based on these disturbances is calculated as explained in Section~\ref{sec:definition}.

\section{Experimental Setup}
\label{sec:ExperimentalSetup}

To yield a comprehensive analysis of the robustness score and the aforementioned testing framework, we included a wide range of CPS datasets as well as many commonly used forecasting models.
The datasets used in this study include:
\begin{itemize}
    \item \textbf{Electricity}\cite{wuAutoformerDecompositionTransformers2021}: This dataset provides hourly electricity consumption records for 321 customers over the period from 2012 to 2014.
    \item \textbf{ETT}\cite{zhouInformerEfficientTransformer2021}: The ETT dataset captures measurements from electronic transformers, including load and oil temperature, over a two-year period.
    It is offered at two temporal resolutions:
    Hourly aggregated data with 17,420 observations across 7 features (ETTh1) and data recorded at 15-minute intervals, comprising 69,680 observations across 7 features (ETTm1).
    \item \textbf{SKAB}\cite{katserSKABSkoltechAnomaly2020}: The SKAB dataset contains sensor readings from a water circulation system testbed.
    Data were collected at a one-second sampling rate over a three-hour period, resulting in 9,405 observations of 8 sensor variables.
    \item \textbf{SWaT}\cite{gohDatasetSupportResearch2017}: The Secure Water Treatment (SWaT) dataset by iTrust, Centre for Research in Cyber Security, Singapore University of Technology and Design consists of sensor and actuator measurements obtained from a scaled-down version of an industrial water treatment plant.
    Data were recorded at one-second intervals continuously over seven days, yielding 449,919 observations with 52 features.
    \item \textbf{Water Quality}\cite{rehbachGECCO2018Industrial2018}: This dataset monitors water quality in the water supply system serving a state in Germany, with 9 sensor readings taken at one-minute intervals from August to November 2016.
\end{itemize}

In this work, we investigate a diverse set of forecasting models spanning several architectural families, including relatively simple fully connected neural networks, recurrent neural networks (RNN), convolutional neural networks (CNN), state-space models (SSM) and Transformers.
In particular, our study considers the following models:

\begin{itemize}
    \item \textbf{DLinear}\cite{zengAreTransformersEffective2023}: A linear forecasting model that integrates a time series decomposition strategy to separately model trend and seasonal components.
    \item \textbf{MLP}\cite{rumelhartLearningRepresentationsBackpropagating1986}: A Multi-Layer Perceptron that employs a series of linear transformations interleaved with ReLU activation functions \cite{nairRectifiedLinearUnits2010} and batch normalization \cite{ioffeBatchNormalizationAccelerating2015} to enhance convergence and generalization.
    \item \textbf{TCN}\cite{leaTemporalConvolutionalNetworks2017}: A Temporal Convolutional Network that leverages causal convolutions and dilated convolutions to capture long-range dependencies of temporal data.
    \item \textbf{LSTM}\cite{hochreiterLongShortTermMemory1997}: A Long Short-Term Memory network that incorporates gating mechanisms to mitigate the vanishing gradient problem.
    \item \textbf{GRU}\cite{choLearningPhraseRepresentations2014}: A Gated Recurrent Unit network that simplifies the LSTM architecture by combining the forget and input gates into a single update gate.
    \item \textbf{RIMs}\cite{goyalRecurrentIndependentMechanisms2021}: A Recurrent Independent Mechanisms model that decomposes the hidden state into multiple independent modules.
    Each module selectively processes different aspects of the input, thereby enhancing robustness.
    \item \textbf{Transformer}\cite{vaswaniAttentionAllYou2017}: A self-attention based architecture that abandons recurrence in favor of parallelized computations.
    \item \textbf{Informer}\cite{zhouInformerEfficientTransformer2021}: An efficient variant of the Transformer specifically designed for long sequence forecasting.
    \item \textbf{Mamba}\cite{guMambaLinearTimeSequence2024}: A SSM that exploits linear time-invariant dynamics to efficiently model sequential data.    
\end{itemize}

Further information about the experimental setup include:

\paragraph{Objective and Hyperparameter Tuning:} 
Our training objective is to forecast the next 30 time steps given the previous 90 time steps, with all models optimized to minimize the MSE using backpropagation.
Extensive hyperparameter tuning is performed via grid search on each dataset separately.
In total, over 10.000 models across 9 model architectures and 6 datasets have been trained.
The specific hyperparameter and their respective ranges are available in our  \hyperlink{https://github.com/awindmann/cps-robustness-benchmark}{online repository}.

\paragraph{Data Splitting and Preprocessing:} 
Each dataset is partitioned into fixed time splits, allocating 70\% for training, 15\% for validation, and 15\% for testing.
To mitigate potential data leakage, we insert a purging gap equivalent to 1\% of the data between each split.
The training data is standardized using its own statistical properties, and the same transformation is applied to the validation and test sets.

\paragraph{Optimization and Batching:} 
All models are trained using a batch size of 64.
The fixed learning rate is chosen as part of the hyperparameter tuning.
The optimization is carried out using the Adam optimizer \cite{DBLP:journals/corr/KingmaB14}.

\paragraph{Early Stopping and Model Checkpointing:} 
Early stopping is employed with a patience of 5 epochs based on the validation loss.
The checkpoint corresponding to the lowest MSE on the validation set is retained as the final model for each architecture and dataset.

\paragraph{Computational Resources:}
All training is executed on a Kubernetes cluster comprising three nodes, each equipped with 128 CPU cores and 500GB of memory, alongside a total of 4 NVIDIA A30 GPUs.
Similar to the approach presented in \cite{steudeEndtoendMLOpsIntegration2024a}, we orchestrated the execution of the experiments using Kubeflow pipelines.

\paragraph{Reproducibility:} 
A fixed random seed is set across all individual model trainings to ensure reproducibility.
Furthermore, the entire framework and experiment configuration is documented and publicly available in our \hyperlink{https://github.com/awindmann/cps-robustness-benchmark}{online repository}.

\section{Results}
\label{sec:results}
In this section, we present the forecasting performance of the evaluated models across multiple real-world CPS datasets, quantify their robustness using the robustness score (Eq.~\eqref{eq:R-score-global}), and further analyze the specific disturbance scenarios.

\subsection{Baseline Forecasting Performance}

Table~\ref{tab:forecasting_performance_baseline} summarizes the forecasting performance, reporting mean and standard deviation of the MSE for validation and test sets across six standardized datasets.
DLinear achieves the best overall forecasting performance, closely followed by Transformer-based models.
Notably, a simple MLP outperforms dedicated time-series models such as TCN, LSTM, GRU and Mamba, indicating the efficacy of simpler architectures.
The Mamba model performs poorly and exhibits considerable variability across test datasets, highlighting its sensitivity to dataset-specific features.

Additionally, we observe the generalization capabilities of the models, reflected by their performance consistency between validation and test datasets.
DLinear exhibits minimal variation between validation and test performance, indicating good generalization.
Transformer variants, MLP, and RIMs display moderate generalization capabilities with slightly larger performance drops.
In contrast, models such as TCN and LSTM experience more significant reductions, suggesting weaker generalization.

\begin{table}[b]
\setlength{\tabcolsep}{4pt}
\centering
\caption{Forecasting performance by model architecture. Values are the mean $\pm$ standard deviation across six datasets; lower values indicate better performance. Best performance values are shown in bold and second-best values are underlined.}
\begin{tabular}{llcc}
\toprule
\textbf{Architecture} & \textbf{Model} & \textbf{Validation MSE} & \textbf{Test MSE} \\
\midrule
Fully Connected & DLinear \cite{zengAreTransformersEffective2023} & \textbf{0.2174} $\pm$ 0.1393 & \textbf{0.2587} $\pm$ 0.1809 \\
Fully Connected & MLP \cite{rumelhartLearningRepresentationsBackpropagating1986} & 0.2284 $\pm$ 0.1523 & 0.3916 $\pm$ 0.2460 \\
Convolution     & TCN \cite{leaTemporalConvolutionalNetworks2017}            & 0.2470 $\pm$ 0.1550 & 0.4362 $\pm$ 0.2834 \\
Recurrent       & LSTM \cite{hochreiterLongShortTermMemory1997}              & 0.2575 $\pm$ 0.1758 & 0.5016 $\pm$ 0.3676 \\
Recurrent       & GRU \cite{choLearningPhraseRepresentations2014}             & 0.2393 $\pm$ 0.1679 & 0.4620 $\pm$ 0.3209 \\
Modular         & RIMs \cite{goyalRecurrentIndependentMechanisms2021}           & 0.2383 $\pm$ 0.1655 & 0.3868 $\pm$ 0.2250 \\
Attention       & Transf. \cite{vaswaniAttentionAllYou2017}                  & \underline{0.2200} $\pm$ 0.1575 & 0.3576 $\pm$ 0.2123 \\
Attention       & Informer \cite{zhouInformerEfficientTransformer2021}         & 0.2203 $\pm$ 0.1550 & \underline{0.3552} $\pm$ 0.2006 \\
SSM             & Mamba \cite{guMambaLinearTimeSequence2024}                   & 0.3047 $\pm$ 0.1945 & 0.7177 $\pm$ 0.7299 \\
\bottomrule
\end{tabular}
\label{tab:forecasting_performance_baseline}
\end{table}

\subsection{Robustness Evaluation}

The robustness scores based on MSE are summarized in Table~\ref{tab:robustness_scores}.
Despite demonstrating strong forecasting performance, DLinear exhibits the lowest robustness score, indicating high susceptibility to disturbances.
Recurrent architectures (LSTM, GRU, and the modular RIMs) achieve the highest robustness scores, with LSTM performing best overall.
These findings imply inherent robustness characteristics associated with recurrent-based architectures.

None of the evaluated models attain robustness scores approaching the ideal value of 1 or above, highlighting the challenging nature of the proposed benchmark framework.
Furthermore, the substantial variation in robustness scores across datasets emphasizes the necessity of multi-dataset evaluations to reliably quantify and compare model robustness.

\begin{table}[]
\centering
\caption{Robustness scores based on MSE by model architecture. Values are the mean $\pm$ standard deviation across six datasets; higher scores indicate greater robustness. Best performance is shown in bold and the second-best value is underlined.}
\begin{tabular}{llc}
\toprule
\textbf{Architecture} & \textbf{Model} & \textbf{Robustness Score} \\
\midrule
Fully Connected  & DLinear \cite{zengAreTransformersEffective2023}       & 0.2754 $\pm$ 0.2455 \\
Fully Connected  & MLP \cite{rumelhartLearningRepresentationsBackpropagating1986}    & 0.4577 $\pm$ 0.2494 \\
Convolution      & TCN \cite{leaTemporalConvolutionalNetworks2017}            & 0.4498 $\pm$ 0.2814 \\
Recurrent        & LSTM \cite{hochreiterLongShortTermMemory1997}              & \textbf{0.6411} $\pm$ 0.3412 \\
Recurrent        & GRU \cite{choLearningPhraseRepresentations2014}             & \underline{0.5948} $\pm$ 0.3543 \\
Modular          & RIMs \cite{goyalRecurrentIndependentMechanisms2021}           & 0.5886 $\pm$ 0.2856 \\
Attention        & Transf. \cite{vaswaniAttentionAllYou2017}                  & 0.4745 $\pm$ 0.2577 \\
Attention        & Informer \cite{zhouInformerEfficientTransformer2021}         & 0.4748 $\pm$ 0.2670 \\
SSM              & Mamba \cite{guMambaLinearTimeSequence2024}                   & 0.5186 $\pm$ 0.3804 \\
\bottomrule
\end{tabular}
\label{tab:robustness_scores}
\end{table}

\subsection{Detailed Scenario Analysis}

\begin{figure}[b]
    \centering
    \includegraphics[width=.7\columnwidth]{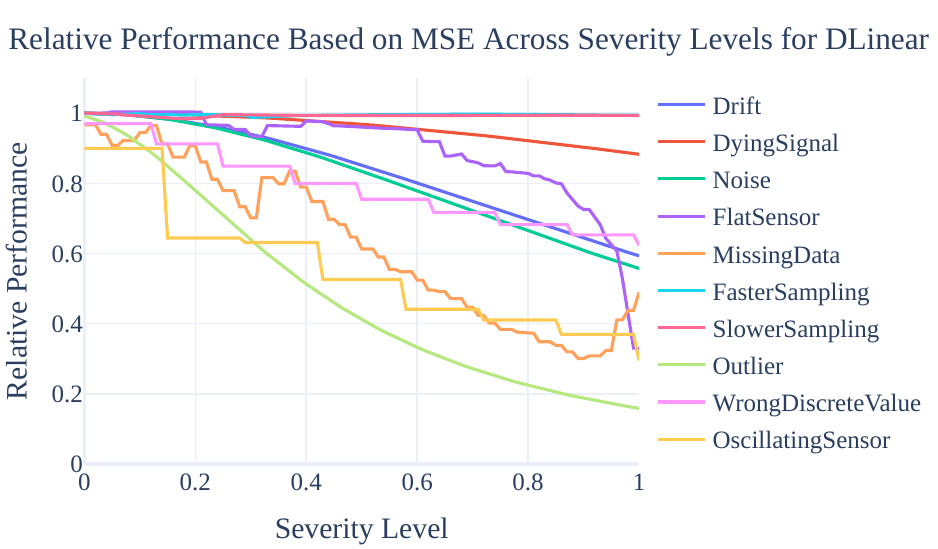}
    \caption{Relative performance of DLinear on the SWaT dataset for multiple test scenarios across severity levels, computed from MSE (Eq.~\eqref{eq:mu-rel}).}
    \label{fig:rel_perf_one_model}
\end{figure}

As described in Section~\ref{sec:definition}, each disturbance was evaluated with increasing severity to assess model susceptibility comprehensively, which is exemplary illustrated for DLinear on SWaT in Figure~\ref{fig:rel_perf_one_model}.
The results show scenario-specific sensitivities: scenarios such as FasterSampling and SlowerSampling have no negative impacts, whereas OscillatingSensor and Outlier notably degrade performance.
Generally, increased severity corresponds to decreased relative performance.
However, in the MissingData scenario, removing more time steps occasionally improved performance at certain severity levels, reflecting the periodic characteristics of CPS data.

\begin{figure*}[]
    \centering
    \includegraphics[width=.99\textwidth]{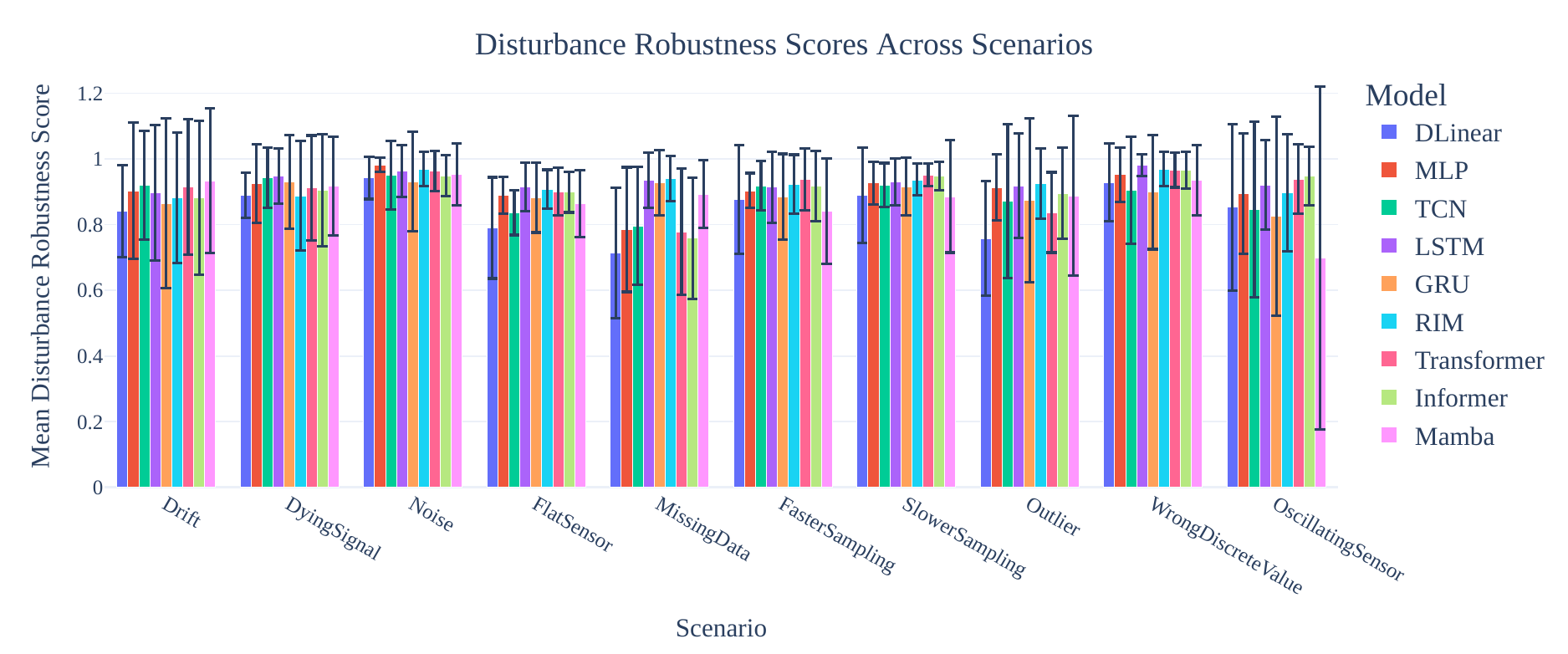}
    \caption{
    Grouped bar chart of the disturbance robustness scores based on the MSE for various models across testing scenarios.
    For each model/scenario combination, the disturbance robustness score (defined in Eq. \eqref{eq:R-score-dist}) is computed and then averaged across six datasets.
    Error bars denote one standard deviation of the dataset-level averages.
    Details about the testing scenarios are provided in Fig.~\ref{fig:scenarios}.
}
    \label{fig:scenario_bar_plot}
\end{figure*}

Figure~\ref{fig:scenario_bar_plot} provides a comprehensive comparison of the models' robustness across individual disturbance scenarios, averaged across all severity levels and datasets.
Here, the disturbance robustness score (see Eq.~\eqref{eq:R-score-dist}) quantifies the models' robustness to individual scenarios.
Most scenarios demonstrate moderate negative impacts with substantial variability across datasets.
The MissingData scenario consistently had the greatest adverse effect overall, with notable architectural differences: recurrent models (LSTM, GRU, RIMs) and Mamba exhibited higher robustness compared to other architectures.
Nevertheless, inter-architecture differences for each disturbance scenario remain relatively modest when considering the high variability across datasets.

\section{Discussion}
\label{sec:discussion}

In this paper, we have introduced a practical definition of robustness tailored specifically to industrial CPS time series and evaluated this definition through a comprehensive forecasting benchmark involving multiple DL architectures.
One insight is the strong performance of relatively simple models such as DLinear, confirming earlier findings \cite{zengAreTransformersEffective2023}.
However, despite its excellent predictive accuracy on both validation and test datasets, DLinear was highly vulnerable to disturbances, achieving the lowest robustness score among all tested architectures.
This shows that standard evaluation methods might not detect overfitting and fail to quantify robustness, which is further affirmed by empirical research in the deployment of machine learning models \cite{paleyesChallengesDeployingMachine2022}.

Overall, recurrent architectures (LSTM, GRU, and modular RIMs) demonstrated high robustness scores.
This could be attributed to their design, which inherently emphasizes recent temporal dependencies, enabling them to better withstand disturbances affecting earlier time steps.
In particular, LSTM exhibited the highest robustness score, potentially explaining its widespread adoption in CPS deployments despite its moderate forecasting performance.
The Transformer-based models displayed strong forecasting performance and moderate robustness, suggesting they could be a suitable compromise between robustness and prognostic accuracy.
Surprisingly, the recently developed Mamba architecture performed poorly both in forecasting accuracy and robustness. 

Robustness scores varied markedly across datasets, highlighting how difficult it is to craft disturbance scenarios that are simultaneously realistic and challenging.
Our present benchmark inevitably leaves several aspects open: it targets only forecasting (not anomaly detection or control) and injects sensor-level faults individually.
Hence, correlated system-wide failures, actuator-side or adversarial attacks remain outside our scope, as do CPS domains such as robotics.

Future work should therefore broaden both model and scenario coverage: adding interpretable statistical or physics-informed baselines, graph-based architectures, large pretrained foundation models and incorporating disturbance-aware training or adversarial augmentation.

\section{Conclusion}
\label{sec:conclusion}
This paper introduced a definition of robustness for prognostic models in CPS and a benchmarking framework to evaluate their robustness under disturbance scenarios.
Our experiments show that while simpler linear models often achieve higher baseline predictive accuracy than complex deep learning models, they are less robust.
Recurrent neural networks offer improved robustness with moderate forecasting performance, whereas Transformers provide a balanced compromise between accuracy and robustness.
These results underscore the importance of jointly considering robustness and forecasting performance during model selection for CPS applications.

\bibliographystyle{unsrt}  
\bibliography{references.bib}  

@inproceedings{bastaniMeasuringNeuralNet2016,
  title = {Measuring {{Neural Net Robustness}} with {{Constraints}}},
  booktitle = {Advances in {{Neural Information Processing Systems}}},
  author = {Bastani, Osbert and Ioannou, Yani and Lampropoulos, Leonidas and Vytiniotis, Dimitrios and Nori, Aditya and Criminisi, Antonio},
  year = {2016},
  doi =  {10.48550/arXiv.1605.07262},
  volume = {29},
  publisher = {Curran Associates, Inc.},
  urldate = {2024-06-28}
}

@book{billintonReliabilityEvaluationEngineering1992,
  title = {Reliability Evaluation of Engineering Systems: Concepts and Techniques},
  shorttitle = {Reliability Evaluation of Engineering Systems},
  author = {Billinton, Roy and Allan, Ronald N.},
  year = {1992},
  edition = {2nd ed},
  publisher = {Plenum Press},
  address = {New York},
  isbn = {978-0-306-44063-2},
  lccn = {TA169 .B54 1992}
}

@article{buzhinskyMetricsMethodsRobustness2021,
  title = {Metrics and Methods for Robustness Evaluation of Neural Networks with Generative Models},
  author = {Buzhinsky, Igor and Nerinovsky, Arseny and Tripakis, Stavros},
  year = {2021},
  month = jul,
  journal = {Machine Learning},
  doi = {10.1007/s10994-021-05994-9},
  urldate = {2023-01-19},
  volume ={112}, 
  pages = {3977--4012},
  langid = {english}
}

@inproceedings{chengRobustTSFTheoryDesign2024,
  title = {{{RobustTSF}}: {{Towards Theory}} and {{Design}} of {{Robust Time Series Forecasting}} with {{Anomalies}}},
  shorttitle = {{{RobustTSF}}},
  booktitle = {Proceedings of the 12th {{International Conference}} on {{Learning Representations}}},
  author = {Cheng, Hao and Wen, Qingsong and Liu, Yang and Sun, Liang},
  year = {2024},
  month = feb,
  doi = {10.48550/arXiv.2402.02032},
  urldate = {2025-03-12}
}

@inproceedings{choLearningPhraseRepresentations2014,
  title = {Learning {{Phrase Representations}} Using {{RNN Encoder}}--{{Decoder}} for {{Statistical Machine Translation}}},
  booktitle = {Proceedings of the 2014 {{Conference}} on {{Empirical Methods}} in {{Natural Language Processing}} ({{EMNLP}})},
  author = {Cho, Kyunghyun and {van Merri{\"e}nboer}, Bart and Gulcehre, Caglar and Bahdanau, Dzmitry and Bougares, Fethi and Schwenk, Holger and Bengio, Yoshua},
  year = {2014},
  month = oct,
  pages = {1724--1734},
  publisher = {Association for Computational Linguistics},
  address = {Doha, Qatar},
  doi = {10.3115/v1/D14-1179},
  urldate = {2023-05-15}
}

@inproceedings{DBLP:journals/corr/KingmaB14,
  title = {Adam: {{A}} Method for Stochastic Optimization},
  booktitle = {3rd International Conference on Learning Representations, {{ICLR}} 2015, San Diego, {{CA}}, {{USA}}, May 7-9, 2015, Conference Track Proceedings},
  author = {Kingma, Diederik P. and Ba, Jimmy},
  editor = {Bengio, Yoshua and LeCun, Yann},
  year = {2015},
  doi = {10.48550/arXiv.1412.6980},
  bibsource = {dblp computer science bibliography, https://dblp.org}
}

@misc{DINSPEC9200112019,
  title = {{{DIN SPEC}} 92001-1:2019-04, {{K{\"u}nstliche Intelligenz}}\_- {{Life Cycle Prozesse}} Und {{Qualit{\"a}tsanforderungen}}\_- {{Teil}}\_1: {{Qualit{\"a}ts-Meta-Modell}}; {{Text Englisch}}},
  shorttitle = {{{DIN SPEC}} 92001-1},
  year = {2019},
  month = apr,
  publisher = {Beuth Verlag GmbH},
  doi = {10.31030/3050203},
  urldate = {2023-11-30}
}

@inproceedings{dixMeasuringRobustnessML2023,
  title = {Measuring the {{Robustness}} of {{ML Models Against Data Quality Issues}} in {{Industrial Time Series Data}}},
  booktitle = {2023 {{IEEE}} 21st {{International Conference}} on {{Industrial Informatics}} ({{INDIN}})},
  author = {Dix, Marcel and Manca, Gianluca and Okafor, Kenneth Chigozie and Borrison, Reuben and Kirchheim, Konstantin and Sharma, Divyasheel and Chandrika, Kr and Maduskar, Deepti and Ortmeier, Frank},
  year = {2023},
  month = jul,
  pages = {1--8},
  publisher = {IEEE},
  address = {Lemgo, Germany},
  doi = {10.1109/INDIN51400.2023.10218129},
  urldate = {2023-12-13}
}

@inproceedings{dziugaiteSearchRobustMeasures2020,
  title = {In Search of Robust Measures of Generalization},
  booktitle = {Advances in {{Neural Information Processing Systems}}},
  author = {Dziugaite, Gintare Karolina and Drouin, Alexandre and Neal, Brady and Rajkumar, Nitarshan and Caballero, Ethan and Wang, Linbo and Mitliagkas, Ioannis and Roy, Daniel M},
  year = {2020},
  volume = {33},
  doi = {10.48550/arXiv.2010.11924},
  pages = {11723--11733},
  publisher = {Curran Associates, Inc.},
  urldate = {2024-10-11}
}

@inproceedings{gohDatasetSupportResearch2017,
  title = {A {{Dataset}} to {{Support Research}} in the {{Design}} of {{Secure Water Treatment Systems}}},
  booktitle = {Critical {{Information Infrastructures Security}}},
  author = {Goh, Jonathan and Adepu, Sridhar and Junejo, Khurum Nazir and Mathur, Aditya},
  editor = {Havarneanu, Grigore and Setola, Roberto and Nassopoulos, Hypatia and Wolthusen, Stephen},
  year = {2017},
  series = {Lecture {{Notes}} in {{Computer Science}}},
  pages = {88--99},
  publisher = {Springer International Publishing},
  address = {Cham},
  doi = {10.1007/978-3-319-71368-7_8},
  langid = {english}
}

@inproceedings{goyalRecurrentIndependentMechanisms2021,
  title = {Recurrent {{Independent Mechanisms}}},
  booktitle = {9th {{International Conference}} on {{Learning Representations}}, {{ICLR}} 2021, {{Virtual Event}}, {{Austria}}, {{May}} 3-7, 2021},
  author = {Goyal, Anirudh and Lamb, Alex and Hoffmann, Jordan and Sodhani, Shagun and Levine, Sergey and Bengio, Yoshua and Sch{\"o}lkopf, Bernhard},
  year = {2021},
  doi = {10.48550/arXiv.1909.10893},
  publisher = {OpenReview.net},
  urldate = {2023-04-28}
}

@inproceedings{guMambaLinearTimeSequence2024,
  title = {Mamba: {{Linear-Time Sequence Modeling}} with {{Selective State Spaces}}},
  shorttitle = {Mamba},
  booktitle = {First {{Conference}} on {{Language Modeling}}},
  author = {Gu, Albert and Dao, Tri},
  year = {2024},
  month = aug,
  doi = {10.48550/arXiv.2312.00752},
  urldate = {2024-12-30},
  langid = {english}
}

@inproceedings{hendrycksManyFacesRobustness2021,
  title = {The {{Many Faces}} of {{Robustness}}: {{A Critical Analysis}} of {{Out-of-Distribution Generalization}}},
  shorttitle = {The {{Many Faces}} of {{Robustness}}},
  booktitle = {2021 {{IEEE}}/{{CVF International Conference}} on {{Computer Vision}} ({{ICCV}})},
  author = {Hendrycks, Dan and Basart, Steven and Mu, Norman and Kadavath, Saurav and Wang, Frank and Dorundo, Evan and Desai, Rahul and Zhu, Tyler and Parajuli, Samyak and Guo, Mike and Song, Dawn and Steinhardt, Jacob and Gilmer, Justin},
  year = {2021},
  month = oct,
  pages = {8320--8329},
  publisher = {IEEE},
  address = {Montreal, QC, Canada},
  doi = {10.1109/ICCV48922.2021.00823},
  urldate = {2025-02-14},
  copyright = {https://doi.org/10.15223/policy-029}
}

@misc{hendrycksUnsolvedProblemsML2022,
  title = {Unsolved {{Problems}} in {{ML Safety}}},
  author = {Hendrycks, Dan and Carlini, Nicholas and Schulman, John and Steinhardt, Jacob},
  year = {2022},
  month = jun,
  number = {arXiv:2109.13916},
  eprint = {2109.13916},
  primaryclass = {cs},
  publisher = {arXiv},
  doi = {10.48550/arXiv.2109.13916},
  urldate = {2025-02-14},
  archiveprefix = {arXiv}
}

@misc{hochreiterLongShortTermMemory1997,
  author = {Hochreiter, Sepp and Schmidhuber, J\"{u}rgen},
  title = {Long Short-Term Memory},
  year = {1997},
  issue_date = {November 15, 1997},
  publisher = {MIT Press},
  address = {Cambridge, MA, USA},
  volume = {9},
  number = {8},
  url = {https://doi.org/10.1162/neco.1997.9.8.1735},
  doi = {10.1162/neco.1997.9.8.1735},
}

@inproceedings{ioffeBatchNormalizationAccelerating2015,
  title = {Batch Normalization: Accelerating Deep Network Training by Reducing Internal Covariate Shift},
  shorttitle = {Batch Normalization},
  booktitle = {Proceedings of the 32nd {{International Conference}} on {{International Conference}} on {{Machine Learning}} - {{Volume}} 37},
  author = {Ioffe, Sergey and Szegedy, Christian},
  year = {2015},
  month = jul,
  series = {{{ICML}}'15},
  pages = {448--456},
  publisher = {JMLR.org},
  address = {Lille, France},
  urldate = {2025-02-04},
  doi = {10.48550/arXiv.1502.03167}
}

@misc{ISOIECTR2021,
  title = {{{ISO}}/{{IEC TR}} 24029-1:2021 --- {{Artificial Intelligence}} ({{AI}}) --- {{Assessment}} of the Robustness of Neural Networks --- {{Part}} 1: {{Overview}}},
  shorttitle = {{{ISO}}/{{IEC TR}} 24029-1},
  year = {2021},
  urldate = {2024-01-12},
  langid = {english}
}

@inproceedings{jiangFantasticGeneralizationMeasures2019,
  title = {Fantastic {{Generalization Measures}} and {{Where}} to {{Find Them}}},
  booktitle = {International {{Conference}} on {{Learning Representations}}},
  author = {Jiang, Yiding and Neyshabur, Behnam and Mobahi, Hossein and Krishnan, Dilip and Bengio, Samy},
  year = {2019},
  month = sep,
  urldate = {2024-10-16},
  doi = {10.48550/arXiv.1912.02178},
  langid = {english}
}

@misc{katserSKABSkoltechAnomaly2020,
  title = {{{SKAB}} - {{Skoltech Anomaly Benchmark}}},
  author = {Katser, Iurii D. and Kozitsin, Vyacheslav O.},
  year = {2020},
  publisher = {Kaggle},
  doi = {10.34740/KAGGLE/DSV/1693952},
  urldate = {2025-02-03},
  langid = {english}
}

@inproceedings{leaTemporalConvolutionalNetworks2017,
  title = {Temporal {{Convolutional Networks}} for {{Action Segmentation}} and {{Detection}}},
  booktitle = {Proceedings of the {{IEEE Conference}} on {{Computer Vision}} and {{Pattern Recognition}}},
  author = {Lea, Colin and Flynn, Michael D. and Vidal, Rene and Reiter, Austin and Hager, Gregory D.},
  year = {2017},
  doi = {https://doi.org/10.48550/arXiv.1611.05267},
  pages = {156--165},
  urldate = {2023-04-16}
}

@article{leePrognosticsHealthManagement2014,
  title = {Prognostics and Health Management Design for Rotary Machinery Systems---{{Reviews}}, Methodology and Applications},
  author = {Lee, Jay and Wu, Fangji and Zhao, Wenyu and Ghaffari, Masoud and Liao, Linxia and Siegel, David},
  year = {2014},
  month = jan,
  journal = {Mechanical Systems and Signal Processing},
  volume = {42},
  number = {1},
  pages = {314--334},
  doi = {10.1016/j.ymssp.2013.06.004},
  urldate = {2023-05-10},
  langid = {english}
}

@misc{mohammedEffectsDataQuality2024,
  title = {The {{Effects}} of {{Data Quality}} on {{Machine Learning Performance}}},
  author = {Mohammed, Sedir and Budach, Lukas and Feuerpfeil, Moritz and Ihde, Nina and Nathansen, Andrea and Noack, Nele and Patzlaff, Hendrik and Naumann, Felix and Harmouch, Hazar},
  year = {2024},
  journal = {Information Systems Information Systems},
  volume = {132},
  pages = {102549},
  number = {arXiv:2207.14529},
  eprint = {2207.14529},
  primaryclass = {cs},
  publisher = {arXiv},
  doi = {10.48550/arXiv.2207.14529},
  urldate = {2023-12-17},
  archiveprefix = {arXiv}
}

@article{moskalenkoResilienceResilientSystems2023,
  title = {Resilience and {{Resilient Systems}} of {{Artificial Intelligence}}: {{Taxonomy}}, {{Models}} and {{Methods}}},
  shorttitle = {Resilience and {{Resilient Systems}} of {{Artificial Intelligence}}},
  author = {Moskalenko, Viacheslav and Kharchenko, Vyacheslav and Moskalenko, Alona and Kuzikov, Borys},
  year = {2023},
  month = mar,
  journal = {Algorithms},
  volume = {16},
  number = {3},
  pages = {165},
  publisher = {Multidisciplinary Digital Publishing Institute},
  doi = {10.3390/a16030165},
  urldate = {2023-12-20},
  copyright = {http://creativecommons.org/licenses/by/3.0/},
  langid = {english}
}

@inproceedings{nairRectifiedLinearUnits2010,
  title = {Rectified Linear Units Improve Restricted Boltzmann Machines},
  booktitle = {Proceedings of the 27th {{International Conference}} on {{International Conference}} on {{Machine Learning}}},
  author = {Nair, Vinod and Hinton, Geoffrey E.},
  year = {2010},
  month = jun,
  series = {{{ICML}}'10},
  pages = {807--814},
  publisher = {Omnipress},
  address = {Madison, WI, USA},
  urldate = {2025-02-04},
  doi = {10.5555/3104322.3104425}
}

@article{paleyesChallengesDeployingMachine2022,
  title = {Challenges in {{Deploying Machine Learning}}: {{A Survey}} of {{Case Studies}}},
  shorttitle = {Challenges in {{Deploying Machine Learning}}},
  author = {Paleyes, Andrei and Urma, Raoul-Gabriel and Lawrence, Neil D.},
  year = {2022},
  journal = {ACM Computing Surveys},
  volume = {55},
  number = {6},
  pages = {1--29},
  doi = {10.1145/3533378},
  urldate = {2023-08-02},
  langid = {english}
}

@article{perez-cerrolazaArtificialIntelligenceSafetyCritical2023,
  title = {Artificial {{Intelligence}} for {{Safety-Critical Systems}} in {{Industrial}} and {{Transportation Domains}}: {{A Survey}}},
  shorttitle = {Artificial {{Intelligence}} for {{Safety-Critical Systems}} in {{Industrial}} and {{Transportation Domains}}},
  author = {{Perez-Cerrolaza}, Jon and Abella, Jaume and Borg, Markus and Donzella, Carlo and Cerquides, Jes{\'u}s and Cazorla, Francisco J. and Englund, Cristofer and Tauber, Markus and Nikolakopoulos, George and Flores, Jose Luis},
  year = {2023},
  month = oct,
  journal = {ACM Computing Surveys},
  doi = {10.1145/3626314},
  volume = {56},
  number = {7},
  pages = {1--40},
  urldate = {2024-02-14}
}

@inproceedings{pfrommerReduceHandicapPerformance2023,
  title = {Reduce the {{Handicap}}: {{Performance Estimation}} for {{AI Systems Safety Certification}}},
  shorttitle = {Reduce the {{Handicap}}},
  booktitle = {2023 {{IEEE}} 21st {{International Conference}} on {{Industrial Informatics}} ({{INDIN}})},
  author = {Pfrommer, Julius and Poyer, Matthieu and Kiroriwal, Saksham},
  year = {2023},
  month = jul,
  pages = {1--7},
  doi = {10.1109/INDIN51400.2023.10218017},
  urldate = {2024-07-05}
}

@techreport{RegulationEU20242024,
  title        = {{Regulation (EU) 2024/1689 of the European Parliament and of the Council of 13 June 2024 on harmonised rules on fair access to and use of data (Data Act)}},
  institution  = {European Union},
  year         = {2024},
  number       = {EU 2024/1689},
  month        = jun,
  howpublished = {\url{https://eur-lex.europa.eu/eli/reg/2024/1689/oj/eng}},
  note         = {Official Journal of the European Union, L, Vol. 2024, 1689},
}

@misc{rehbachGECCO2018Industrial2018,
  author       = {S. Moritz and F. Rehbach and S. Chandrasekaran and M. Rebolledo and Thomas Bartz-Beielstein},
  title        = {{GECCO Industrial Challenge 2018 Dataset: A water quality dataset for the 'Internet of Things: Online Anomaly Detection for Drinking Water Quality' competition at the Genetic and Evolutionary Computation Conference 2018, Kyoto, Japan}},
  howpublished = {Zenodo},
  year         = {2018},
  month        = feb,
  note         = {doi: \url{10.5281/zenodo.3884398}},
  url          = {10.5281/zenodo.3884398}
}

@article{rumelhartLearningRepresentationsBackpropagating1986,
  title = {Learning Representations by Back-Propagating Errors},
  author = {Rumelhart, David E. and Hinton, Geoffrey E. and Williams, Ronald J.},
  year = {1986},
  month = oct,
  journal = {Nature},
  volume = {323},
  number = {6088},
  pages = {533--536},
  publisher = {Nature Publishing Group},
  doi = {10.1038/323533a0},
  urldate = {2025-02-04},
  copyright = {1986 Springer Nature Limited},
  langid = {english}
}

@article{runggerNotionRobustnessCyberPhysical2016,
  title = {A {{Notion}} of {{Robustness}} for {{Cyber-Physical Systems}}},
  author = {Rungger, Matthias and Tabuada, Paulo},
  year = {2016},
  month = aug,
  journal = {IEEE Transactions on Automatic Control},
  volume = {61},
  number = {8},
  pages = {2108--2123},
  doi = {10.1109/TAC.2015.2492438},
  urldate = {2025-02-11},
  copyright = {https://ieeexplore.ieee.org/Xplorehelp/downloads/license-information/IEEE.html}
}

@inproceedings{schelterJENGAFrameworkStudy2021,
  title = {{{JENGA}} - {{A Framework}} to {{Study}} the {{Impact}} of {{Data Errors}} on the {{Predictions}} of {{Machine Learning Models}}},
  booktitle = {International {{Conference}} on {{Extending Database Technology}}},
  author = {Schelter, Sebastian and Rukat, Tammo and Biessmann, Felix},
  year = {2021},
  publisher = {OpenProceedings.org},
  doi = {10.5441/002/EDBT.2021.63},
  urldate = {2024-06-27},
  langid = {english}
}

@inproceedings{schmidtAssessingRobustnessDataDriven2025,
  title = {Assessing {{Robustness}} in {{Data-Driven Modeling}} of {{Cyber-Physical Systems}}},
  booktitle = {Machine {{Learning}} for {{Cyber-Physical Systems}} ({{ML4CPS}})},
  author = {Schmidt, Maximilian and Plambeck, Swantje and Fey, Goerschwin},
  year = {2025},
  address = {Berlin},
  doi = {10.24405/20020},
  langid = {english}
}

@article{schuettRiskManagementArtificial2024,
  title = {Risk {{Management}} in the {{Artificial Intelligence Act}}},
  author = {Schuett, Jonas},
  year = {2024},
  month = jun,
  journal = {European Journal of Risk Regulation},
  volume = {15},
  number = {2},
  pages = {367--385},
  doi = {10.1017/err.2023.1},
  urldate = {2025-01-06},
  langid = {english}
}

@inproceedings{sinhaCertifyingDistributionalRobustness2018,
  title = {Certifying {{Some Distributional Robustness}} with {{Principled Adversarial Training}}},
  booktitle = {International {{Conference}} on {{Learning Representations}}},
  author = {Sinha, Aman and Namkoong, Hongseok and Duchi, John},
  year = {2018},
  month = feb,
  urldate = {2024-11-05},
  doi = {10.48550/arXiv.1710.10571},
  langid = {english}
}

@inproceedings{steudeEndtoendMLOpsIntegration2024a,
  title = {End-to-End {{MLOps}} Integration: A Case Study with {{ISS}} Telemetry Data},
  shorttitle = {End-to-End {{MLOps}} Integration},
  booktitle = {{{ML4CPS}} -- {{Machine Learning}} for {{Cyber-Physical Systems}}},
  author = {Steude, Henrik Sebastian and Geier, Christian and Moddemann, Lukas and Creutzenberg, Martin and Pfeifer, Jann and Turk, Samo and Niggemann, Oliver},
  year = {2024},
  month = mar,
  publisher = {UB HSU},
  address = {Berlin},
  doi = {10.24405/15316},
  urldate = {2025-02-04},
  collaborator = {{Helmut-Schmidt-Universit{\"a}t}},
  copyright = {open access},
  langid = {english}
}

@inproceedings{szegedyIntriguingPropertiesNeural2014,
  title = {Intriguing Properties of Neural Networks},
  booktitle = {International {{Conference}} on {{Learning Representations}}},
  author = {Szegedy, Christian and Zaremba, Wojciech and Sutskever, Ilya and Bruna, Joan and Erhan, Dumitru and Goodfellow, Ian and Fergus, Rob},
  year = {2014},
  month = feb,
  eprint = {1312.6199},
  primaryclass = {cs},
  publisher = {arXiv},
  doi = {10.48550/arXiv.1312.6199},
  urldate = {2024-02-14},
  archiveprefix = {arXiv}
}

@inproceedings{vaswaniAttentionAllYou2017,
  title = {Attention Is {{All}} You {{Need}}},
  booktitle = {Advances in {{Neural Information Processing Systems}}},
  author = {Vaswani, Ashish and Shazeer, Noam and Parmar, Niki and Uszkoreit, Jakob and Jones, Llion and Gomez, Aidan N and Kaiser, {\L}ukasz and Polosukhin, Illia},
  year = {2017},
  volume = {30},
  doi = {10.48550/arXiv.1706.03762},
  publisher = {Curran Associates, Inc.},
  urldate = {2023-04-28}
}

@inproceedings{windmannArtificialIntelligenceIndustry2024,
  title = {Artificial {{Intelligence}} in {{Industry}} 4.0: {{A Review}} of {{Integration Challenges}} for {{Industrial Systems}}},
  shorttitle = {Artificial {{Intelligence}} in {{Industry}} 4.0},
  booktitle = {2024 {{IEEE}} 22nd {{International Conference}} on {{Industrial Informatics}} ({{INDIN}})},
  author = {Windmann, Alexander and Wittenberg, Philipp and Schieseck, Marvin and Niggemann, Oliver},
  year = {2024},
  month = aug,
  pages = {1--8},
  doi = {10.1109/INDIN58382.2024.10774364},
  urldate = {2025-01-10},
  copyright = {All rights reserved}
}

@inproceedings{windmannRobustnessGeneralizationPerformance2023,
  title = {Robustness and {{Generalization Performance}} of {{Deep Learning Models}} on {{Cyber-Physical Systems}}: {{A Comparative Study}}},
  booktitle = {{{IJCAI}} 2023 {{Workshop}} of {{Artificial Intelligence}} for {{Time Series Analysis}} ({{AI4TS}})},
  author = {Windmann, Alexander and Steude, Henrik and Niggemann, Oliver},
  year = {2023},
  doi = {10.48550/arXiv.2306.07737},
  copyright = {All rights reserved},
  langid = {english}
}

@inproceedings{wuAutoformerDecompositionTransformers2021,
  title = {Autoformer: {{Decomposition Transformers}} with {{Auto-Correlation}} for {{Long-Term Series Forecasting}}},
  shorttitle = {Autoformer},
  booktitle = {Advances in {{Neural Information Processing Systems}}},
  author = {Wu, Haixu and Xu, Jiehui and Wang, Jianmin and Long, Mingsheng},
  year = {2021},
  volume = {34},
  doi = {10.48550/arXiv.2106.13008},
  pages = {22419--22430},
  publisher = {Curran Associates, Inc.},
  urldate = {2025-01-28}
}

@inproceedings{yoonRobustProbabilisticTime2022,
  title = {Robust {{Probabilistic Time Series Forecasting}}},
  booktitle = {Proceedings of {{The}} 25th {{International Conference}} on {{Artificial Intelligence}} and {{Statistics}}},
  author = {Yoon, Taeho and Park, Youngsuk and Ryu, Ernest K. and Wang, Yuyang},
  year = {2022},
  month = may,
  pages = {1336--1358},
  doi = {10.48550/arXiv.2202.11910},
  publisher = {PMLR},
  urldate = {2025-03-13},
  langid = {english}
}

@article{yuanOutDistributionGeneralization2022,
  title = {Towards out of Distribution Generalization for Problems in Mechanics},
  author = {Yuan, Lingxiao and Park, Harold S. and Lejeune, Emma},
  year = {2022},
  month = oct,
  journal = {Computer Methods in Applied Mechanics and Engineering},
  volume = {400},
  pages = {115569},
  doi = {10.1016/j.cma.2022.115569},
  urldate = {2023-12-07}
}

@article{zengAreTransformersEffective2023,
  title = {Are {{Transformers Effective}} for {{Time Series Forecasting}}?},
  author = {Zeng, Ailing and Chen, Muxi and Zhang, Lei and Xu, Qiang},
  year = {2023},
  month = jun,
  journal = {Proceedings of the AAAI Conference on Artificial Intelligence},
  volume = {37},
  number = {9},
  pages = {11121--11128},
  doi = {10.1609/aaai.v37i9.26317},
  urldate = {2025-02-04}
}

@article{zhouInformerEfficientTransformer2021,
  title = {Informer: {{Beyond Efficient Transformer}} for {{Long Sequence Time-Series Forecasting}}},
  shorttitle = {Informer},
  author = {Zhou, Haoyi and Zhang, Shanghang and Peng, Jieqi and Zhang, Shuai and Li, Jianxin and Xiong, Hui and Zhang, Wancai},
  year = {2021},
  month = may,
  journal = {Proceedings of the AAAI Conference on Artificial Intelligence},
  volume = {35},
  number = {12},
  pages = {11106--11115},
  doi = {10.1609/aaai.v35i12.17325},
  urldate = {2025-01-23},
  langid = {english}
}

@inproceedings{zhouRobustnessTestingData2022,
  title = {Robustness {{Testing}} of {{Data}} and {{Knowledge Driven Anomaly Detection}} in {{Cyber-Physical Systems}}},
  booktitle = {2022 52nd {{Annual IEEE}}/{{IFIP International Conference}} on {{Dependable Systems}} and {{Networks Workshops}} ({{DSN-W}})},
  author = {Zhou, Xugui and Kouzel, Maxfield and Alemzadeh, Homa},
  year = {2022},
  month = jun,
  pages = {44--51},
  publisher = {IEEE Computer Society},
  doi = {10.1109/DSN-W54100.2022.00017},
  urldate = {2023-12-18},
  langid = {english}
}

@article{mitschModelPlexVerifiedRuntime2016,
  title = {{{ModelPlex}}: Verified Runtime Validation of Verified Cyber-Physical System Models},
  shorttitle = {{{ModelPlex}}},
  author = {Mitsch, Stefan and Platzer, Andr{\'e}},
  year = {2016},
  month = oct,
  journal = {Formal Methods in System Design},
  volume = {49},
  number = {1},
  pages = {33--74},
  doi = {10.1007/s10703-016-0241-z},
  urldate = {2025-04-01},
  langid = {english}
}

@article{sunPredictingFutureCapacity2024,
  title = {Predicting the {{Future Capacity}} and {{Remaining Useful Life}} of {{Lithium-Ion Batteries Based}} on {{Deep Transfer Learning}}},
  author = {Sun, Chenyu and Lu, Taolin and Li, Qingbo and Liu, Yili and Yang, Wen and Xie, Jingying},
  year = {2024},
  month = sep,
  journal = {Batteries},
  volume = {10},
  number = {9},
  pages = {303},
  publisher = {Multidisciplinary Digital Publishing Institute},
  doi = {10.3390/batteries10090303},
  urldate = {2025-06-20},
  copyright = {http://creativecommons.org/licenses/by/3.0/},
  langid = {english}
}

@inproceedings{wangHealthIndicatorForecasting2020,
  title = {Health {{Indicator Forecasting}} for {{Improving Remaining Useful Life Estimation}}},
  booktitle = {2020 {{IEEE International Conference}} on {{Prognostics}} and {{Health Management}} ({{ICPHM}})},
  author = {Wang, Qiyao and Farahat, Ahmed and Gupta, Chetan and Wang, Haiyan},
  year = {2020},
  month = jun,
  pages = {1--8},
  doi = {10.1109/ICPHM49022.2020.9187047},
  urldate = {2025-06-20}
}


\end{document}